\documentclass[sigconf]{acmart}
\usepackage[T1]{fontenc}

\AtBeginDocument{%
  \providecommand\BibTeX{{%
    \normalfont B\kern-0.5em{\scshape i\kern-0.25em b}\kern-0.8em\TeX}}}

\setcopyright{none}
\copyrightyear{}
\acmYear{}
\acmDOI{}

\acmConference[DSHealth '21]{DSHealth '21: 2021 KDD Workshop on Applied Data Science for Healthcare}{August 16, 2021}{Virtual}
\acmBooktitle{}
\acmPrice{}
\acmISBN{}

\usepackage{amsmath}
\usepackage{marginnote}
\usepackage{multicol}
\usepackage{enumitem}
\usepackage{natbib}
\usepackage{xcolor}
\usepackage{booktabs}
\usepackage{adjustbox}
\usepackage{subcaption}
\usepackage{array}
\begin{document}

\newcommand{\todotip}[2][NA]{{\color{red} $\rhd${[Assignee #1]: #2}}}
\newcommand{\explannote}[2][NA]{{\reversemarginpar \color{blue} $\lhd${[#1 says]: #2}}}
\newcommand{\numproto}{$20$}
\newcommand{\dm}{\texttt{T2DM}}
\newcommand{{\ckd}}{\texttt{CKD}}

\newcolumntype{A}{>{\raggedright}p{0.75\linewidth}}
\newcolumntype{B}{>{\raggedright}p{0.25\linewidth}}

\title{Leveraging Clinical Context for User-Centered Explainability: A Diabetes Use Case}

\newcommand\Mark[1]{\textsuperscript{#1}}
\author{
Shruthi Chari\Mark{1}, Prithwish Chakraborty\Mark{2}, Mohamed Ghalwash\Mark{2}, Oshani Seneviratne\Mark{1}, \\
  Elif Eyigoz\Mark{2}, Daniel Gruen\Mark{1}, Fernando Suarez Saiz\Mark{3}, Ching-hua Chen\Mark{2}, \\
  Pablo Meyer Rojas\Mark{2}, Deborah L. McGuinness\Mark{1}}
\affiliation{
  \institution{
    \Mark{1} Rensselaer Polytechnic Institute (RPI), NY, USA\\
    \Mark{2} Center for Computational Health, IBM Research, NY, USA\\
    \Mark{3} IBM Watson Health, MA, USA
    }
  \city{ }
  \country{ }
  }

\renewcommand{\shortauthors}{Chari, et al.}


\begin{abstract}
  Academic advances of AI models in high-precision domains, like healthcare, need to be made explainable in order to enhance real-world adoption. Our past studies and ongoing interactions indicate that medical experts can use AI systems with greater trust if there are ways to connect the model inferences about patients to explanations that are tied back to the context of use. Specifically, risk prediction is a complex problem of diagnostic and interventional importance to clinicians wherein they consult different sources to make decisions. To enable the adoption of the ever improving AI risk prediction models in practice, we have begun to explore techniques to contextualize such models along three dimensions of interest: \textit{the patients' clinical state, AI predictions about their risk of complications, and algorithmic explanations supporting the predictions}. We validate the importance of these dimensions by implementing a proof-of-concept (POC) in type-2 diabetes ({\dm}) use case where we assess the risk of chronic kidney disease ({\ckd}) - a common {\dm} comorbidity.  Within the POC, we include risk prediction models for CKD, post-hoc explainers of the predictions, and other natural-language modules which operationalize domain knowledge and CPGs to provide context.  With primary care physicians (PCP) as our end-users, we present our initial results and clinician feedback in this paper. Our POC approach covers multiple knowledge sources and clinical scenarios, blends knowledge to explain data and predictions to PCPs, and received an enthusiastic response from our medical expert. 
 %
\end{abstract}




\maketitle

\section{Introduction}

%

Advances in academic usage of machine learning (ML) techniques in healthcare and other critical applications have not  been adopted at a similar pace in the real-world. This has led to a renewed emphasis  \cite{doshi2017accountability,mittelstadt2019explaining,namreport2019,chakraborty2020tutorial} on devising strategies to better explain Artificial Intelligence (AI) models to end-users.
 In clinical settings, we have found from our user studies~\cite{dan2020designing} that clinicians consult different sources of knowledge when making patient-centric decisions and often require other explanation types/presentations (e.g., contrastive, evidence-based explanations). 
  With the aim to infuse clinical knowledge into explanations, and, we focus on a risk-prediction use case in diagnostic and interventional settings.
  We consider AI risk-prediction models (including algorithmic explainers) and aim to enhance clinicians' confidence in using such models by connecting (or contextualizing) the model outputs to multiple sources. These sources include clinical indicators from patient data as well as domain knowledge from biomedical ontologies and authoritative literature such as clinical practice guidelines (CPGs). As a proof-of-concept (POC), we explore techniques to generate explanations around risk prediction models for comorbid complications among type-2 diabetes ({\dm}) patients, with initial focus on chronic kidney disease ({\ckd}).
 



Our method 
builds on both expert feedback and past efforts to leverage clinical domain knowledge for generating explanations within AI assistants. 
Some 
notable and relevant past works include: MYCIN~\cite{shortliffe1974mycin}, where domain literature was encoded as rules and trace-based explanations, which addressed `Why,' `What,' and `How,' were provided for the treatment of infectious diseases; the DESIREE project~\cite{seroussi2018implementing}, where case, experience, and guideline-based knowledge was used to generate insights relevant to patient cases; and a mortality risk prediction effort~\cite{raghu2021learning} of cardiovascular patients, where a probabilistic model was utilized to combine insights from patient features, and
domain knowledge, to ascertain patient conformance to the literature. However, these approaches are either not flexible nor scalable for the ingestion of new domain knowledge~\cite{shortliffe1974mycin, seroussi2018implementing}, or are narrowly focused in their approach to explanations along limited dimensions~\cite{raghu2021learning}. 
We attempt to allow clinicians to probe the supporting evidence systematically and thoroughly while asking holistic 
questions about the supporting evidence (s) 
to understand their patients better.

To 
support the goal of providing user-centered and clinically relevant explanations in the clinicians' context of use,
we identify three contextualization dimensions.
Later (Sec. \ref{sec:results} and \ref{sec:methods}), we present findings from a POC implementation to demonstrate their feasibility and usability:

\noindent $\bullet$ \textbf{Contextualizing the patient} by connecting their clinical history to treatments typically recommended for such patients, according to CPGs.
\\ \noindent $\bullet$ \textbf{Contextualizing risk predictions for the patient} in terms of the prediction's impact on decisions, based on general norms of practice concerning potential complications, as evident from guidelines and other domain knowledge, including medical ontologies.
\\ \noindent $\bullet$  \textbf{Contextualizing details of algorithmic, post-hoc explanations}, such as connecting features that were the most important to other information based on their potential medical significance, such as  through connections to physiological pathways.

\begin{table*}[]
\caption{Results for a selected set of questions from the various modules (risk prediction models, post-hoc explainers, and guideline QA). 
The annotation column contains answer information including: information
sources, the disease aspect (i.e., {\dm}, {\ckd}, or both), and the contextualization dimension captured. ($^*$ Diseases ordered by frequency)}
\label{tab:resultsquestions}

\begin{adjustbox}{width=1\textwidth}
\small
\begin{tabular}{lll}
\hline
\textbf{Question} &
  \textbf{Annotations} &
  \textbf{Answers} \\ \hline
1. Who are the most interesting patients? &
  \begin{tabular}[c]{@{}l@{}}\textbf{Source}: Algorithmic, \textbf{Relevance:} {\ckd} + {\dm}, \\ \textbf{Contextualization:} Post-hoc Explanations\end{tabular} & Refer to summary of patients in Tab. \ref{tab:protosummary}.
   \\ \hline
\begin{tabular}[c]{@{}l@{}}2. Why does the model state a high-risk for {\ckd}?\end{tabular} &
  \begin{tabular}[c]{@{}l@{}}\textbf{Source:} Algorithmic, \textbf{Relevance:} {\ckd}, \\ \textbf{Contextualization:} risk prediction\end{tabular} & Refer to the feature importances in Fig. \ref{fig:shap:summary_proto}
   \\ \hline
3. How will you describe the patient w.r.t diabetes? &
  \begin{tabular}[c]{@{}l@{}}\textbf{Source:} Algorithmic, \textbf{Relevance:} {\dm}, \\ \textbf{Contextualization:} risk prediction\end{tabular} & \begin{tabular}[c]{@{}l@{}} High HbA1C ($\geq 10$). \\ \textbf{Other top conditions$^*$}: Mood Disorders, \\ 
Endocrine; nutritional; and metabolic diseases and immunity disorders | Disorders of \\lipid metabolism
Diseases of the nervous system and sense organs | Conditions associated \\with dizziness or vertigo 
Diseases of the genitourinary system | Other female genital disorders. \end{tabular}
   \\ \hline
3a. What should be done if A1C levels are greater than 10 ? &
  \begin{tabular}[c]{@{}l@{}}\textbf{Source:} Guidelines, \textbf{Relevance:} {\dm}, \\ \textbf{Contextualization:} patient\end{tabular} &
  \begin{tabular}[c]{@{}l@{}}The early introduction of insulin should be considered if there is evidence of ongoing \\ catabolism (weight loss), if symptoms of hyperglycemia are present, or when A1C levels \\ (greater than 10\% {[} 86 mmol/mol {]})  or blood glucose levels (greater than or equal to \\ 300 mg/dL {[}16.7 mmol/L{]}) are very high.\end{tabular} \\ \hline
\begin{tabular}[c]{@{}l@{}}4. What do you know about the viability of GLP-1 RA \\  drugs for the patient?\end{tabular} &
  \begin{tabular}[c]{@{}l@{}}\textbf{Source:} Guidelines, \textbf{Relevance:} {\dm} + {\ckd}, \\ \textbf{Contextualization:} patient + risk prediction\end{tabular} & 
  { \begin{tabular}[c]{@{}l@{}}The guidelines state that drugs in the GLPA-1 RA family have  a neutral effect on \\ cardiovascular events (both ASCVD and HF) and have a beneficial effect to slow \\ progression  of diabetes kidney disease (patient's  {\ckd} risk is found to be \textbf{0.83}), \\ so this drug is likely to be tolerated by this patient who is diagnosed with \\hypertension and genitourinary diseases.\end{tabular}}  \\ \hline
\begin{tabular}[c]{@{}l@{}}5. What drugs to administer for patients with \\ T2D complications?\end{tabular} &
  \begin{tabular}[c]{@{}l@{}}\textbf{Source:} Guidelines, \textbf{Relevance:} {\dm} + {\ckd}, \\ \textbf{Contextualization:} patient + risk prediction\end{tabular} &
  { \begin{tabular}[c]{@{}l@{}}Among patients with type 2 diabetes who have established atherosclerotic\\  cardiovascular disease or indicators of high risk, established kidney disease, \\ or heart failure, a sodium–glucose co-transporter 2 inhibitor or glucagon-like peptide 1\\  receptor agonist with demonstrated cardiovascular disease benefit (Table 9.1, Table 10.3B, \\ Table 10.3C) is recommended as part of the glucose-lowering regimen independent of A1C \\ and in consideration of patient-specific factors (Fig. 9.1 and Section 10).\end{tabular}} \\ \hline
\begin{tabular}[c]{@{}l@{}}6. What is typically done for patients like this \\ who are not meeting treatment goals?\end{tabular} &
  \begin{tabular}[c]{@{}l@{}}\textbf{Source:} Guidelines, \textbf{Relevance:} {\dm}, \\ \textbf{Contextualization:} patient\end{tabular} &
  \begin{tabular}[c]{@{}l@{}}Recommendation for treatment intensification for patients \\ not meeting treatment goals should not be delayed.\end{tabular} \\ \hline
\end{tabular}
\end{adjustbox}
\end{table*}

\section{Diabetes Use Case}
\label{sec:background}
\subsection{Use Case and Patient Cohort} 

Diabetes is one of the top 5 chronic diseases affecting the adult population in the US~\cite{cdcT2D}. 
Diabetes management involves monitoring for and treating 
related comorbid conditions. 
Effective and timely prediction of such conditions can lead to an overall improvement in the quality of care. We focus on {\dm} and use machine learning (ML) models to predict the risk of developing certain {\dm} comorbid complications. We predict the risk of developing {\ckd}, a commonly occurring micro-vascular complication of {\dm}, within 360 days of {\dm} onset. {\ckd} is one of the leading cause of death in the US~\cite{cdcCKD}, with an estimated $37$ million cases in the US (who are mostly undiagnosed) and cost medicare in 2018 $81.1B$ and end stage renal disease an additional $36.6B$. Typically, actions to prevent onset of {\ckd} among {\dm} patients revolve around proper disease control  including close disease monitoring, proper treatment adherence, and patient education. Incorporating accurate risk prediction of {\ckd} in the clinical workflow can lead to more timely actions, potentially delay the onset of {\ckd} and in some cases prevent its progression.

Under this use case, we explore strategies to provide context around interventions for particular patients, explain their {\dm} state and individual risk factors.
We conduct our analysis on claims sub-component of the Limited IBM MarketScan Explorys Claims-EMR Data Set (LCED) covering both administrative claims and EHR data of over $5$ million commercially insured patients between 2013 and 2017.
Medical diagnoses are encoded 
using International Classification of Diseases (ICD) codes. 
We selected only those {\dm} patients (with ICD9 codes 250.*0, 250.*2, 362.0, and ICD10 code E11)  that satisfied the following criteria as our cohort:
\\ \noindent $\bullet$ have two or more visits with {\dm} diagnosis,
\\ \noindent $\bullet$ were enrolled continuously for $12$ months prior to 
{\dm} diagnosis,
\\ \noindent $\bullet$ their number of visits for {\dm}  is greater than those for other forms of diabetes such as \texttt{T1D}, and
\\ \noindent $\bullet$ their age at the initial {\dm} diagnosis is between 19-64 years.

Among {\dm} patients, we use the first diagnosis of chronic kidney disease ({\ckd}) (ICD10 N18 or ICD9 585.*, 403.*) after the initial diagnosis of {\dm} as the outcome to predict. 
At the time of the first {\dm} diagnosis, we predict the risk of the patient developing {\ckd} within $1$ year using Clinical Classifications Software (CCS) codes, age group, and sex as features for the predictive model.

\subsection{Problem Setup}
In our {\dm} - {\ckd} risk prediction use case, we focus on two scenarios where clinicians are likely to ask questions
concerning patients and disease risk: (1) a population health scenario where they are trying to understand their patient populations better, and (2) the point of care setting where they are deciding on interventional strategies. In these scenarios, when clinicians make decisions, they often tie the content back to evidence from guidelines, other domain knowledge sources like physiological disease hierarchies, or their own expertise. We aim to expose some of the relevant information as explanations to a set of questions, structured around the two scenarios, through our POC implementation for the identified contextualization capabilities.

\subsection{Guidelines} \label{sec:guidelines}
We use the American Diabetes Association (ADA) Standards of Medical Care CPG~\cite{american2021introduction}, as evidence-based, expert knowledge for the contextualization dimensions defined earlier. 
ADA CPGs are released annually and each version of the CPG contains individual chapters that deal with different aspects of T2D diagnosis, treatment, and management (e.g., pharmacologic therapies and microvascular complications chapters). Further, 
within each chapter, recommendation groups consist of recommendations with different grade levels, determined by evidence quality (i.e., evidence supported by meta-analysis are assigned a grade-level higher than those supported by expert opinion alone). We converted the HTML version of the guidelines into a JSON that captures the evidence structure, using an HTML parsing library in Python, BeautifulSoup~\footnote{BeautifulSoup: \url{https://www.crummy.com/software/BeautifulSoup/bs4/doc/}}.
As a first pass for utilizing guidelines as domain knowledge within our generated insights that provide contextualization capabilities, we utilize \textit{17} automatically extracted recommendations, and some tabular content, of two chapters that best align with our prediction of chronic kidney disease complications: `Ch. 11: Microvascular Complications and Foot Care: Standards of Medical Care in Diabetes'\footnote{\url{https://care.diabetesjournals.org/content/44/Supplement_1/S111}} and `Ch. 9: Pharmacologic Approaches to Glycemic Treatment'.\footnote{\url{https://care.diabetesjournals.org/content/44/Supplement_1/S151}} 

\section{Contextualization Capabilities} \label{sec:results}
We present the contextualization capabilities of our method to assist clinicians in managing {\dm} patients. We use inferences from risk prediction, post-hoc explanations, and QA models to address a set of clinically relevant questions that can be answered by our methods and fit in the clinical workflow. 
Table \ref{tab:resultsquestions} presents answers along with their source(s) (i.e., whether from algorithmic results alone or guidelines), the relevance of the answer concerning the disease (i.e., {\dm} alone or including the comorbidity used for risk prediction, {\ckd}), and the contextualization dimension (i.e., whether the answer contextualizes the patient, risk prediction or the post-hoc explanation). 

We have attempted to simulate a question-flow, emulating the process that a clinician would use
while managing and identifying {\dm} patients under their care at a population-health level (\textbf{Q. 1}) and subsequently for a  representative high-risk patient in their care (\textbf{Q. 2 - 6}).
In a point-of-care scenario, we delve into a particular patient from the representative high-risk (prototypical) set by first addressing {\dm} progression (\textbf{Q. 2, 3, 3a}), and then presenting guideline evidence on what is typically done for such patients (\textbf{Q. 4 - 6}). While the first set of questions (\textbf{Q. 2, 3})  helps the clinician understand the patient's current clinical state, the second set (\textbf{Q. 3a - 6}) allows the clinician to utilize the model predictions 
for ongoing disease management decisions.

The results used to populate the answers in Table \ref{tab:resultsquestions} are obtained from insights 
generated by the various modules described in Sec. \ref{sec:methods}, including the risk prediction models (\textbf{Q. 2, 4, 5}), the post-hoc explainers (\textbf{Q. 1, 2}), the guideline QA module (\textbf{Q. 3a, 4, 5, 6}), and in some cases, by the application of simple data analysis algorithms on the {\dm} cohort (\textbf{Q. 1, 3}).

\begin{figure*}
    \centering
    \includegraphics[width=\linewidth]{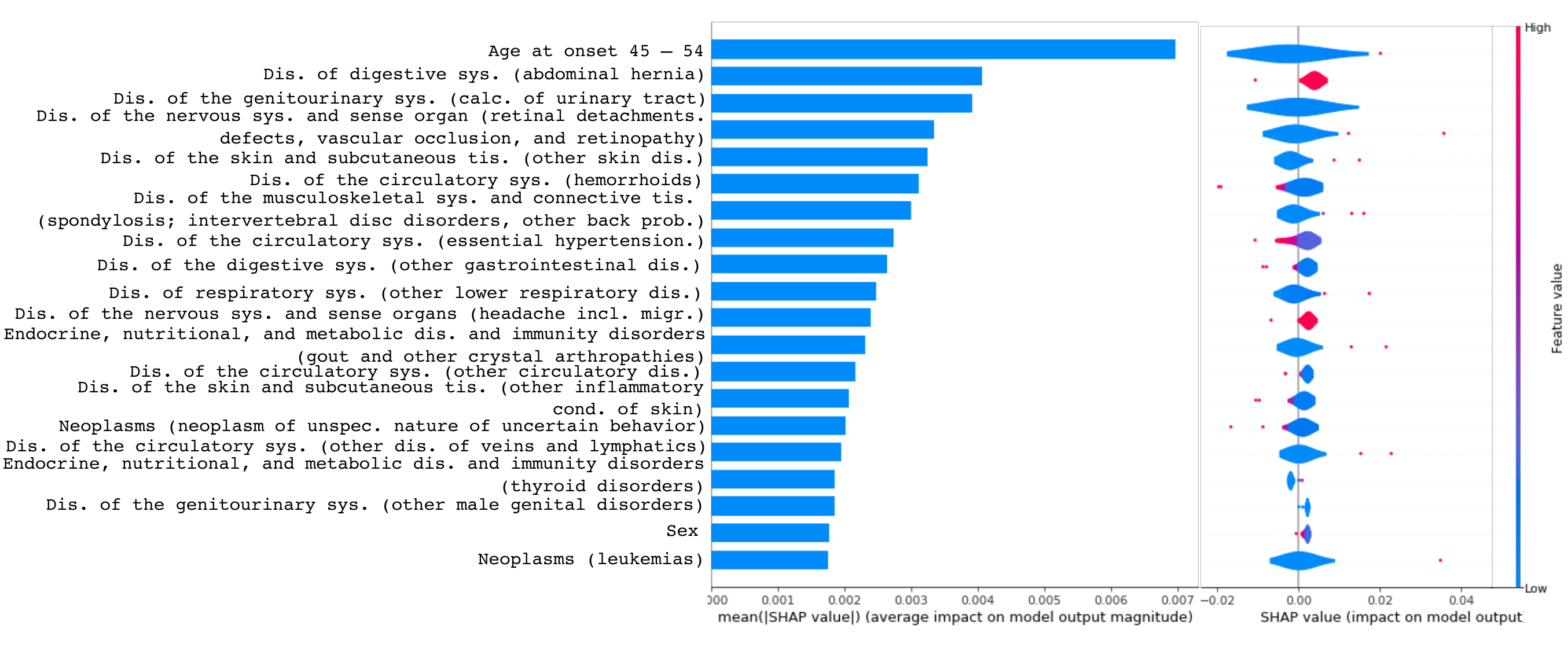}
    \vspace{-2em}
        \caption{Feature importance for {\ckd} prediction among {\numproto} prototypical patients using SHAP (left), showing absolute importance, and (right) showing feature impact on model prediction w.r.t. presence/absence of features}
        \label{fig:shap:summary_proto}
\end{figure*}

Further, to answer questions like {\textbf{Q. 2}}, we contextualize the risk-factors as found by algorithmic explainers. 
Figure~\ref{fig:shap:summary_proto} (left) shows the top $20$ features for the set of {\numproto} prototypical patients under investigation. These prototypical patients are all found to be at high-risk for {\ckd} and hence can be interesting to clinicians within the scope of this {\dm} and {\ckd} use case (\textbf{Q. 1}). For these prototypical patients, we present aggregated feature importances, to account for HIPAA restrictions.  We can see that demographic features, such as age and the presence of other disorders, such as `other skin disorders,' were found to be important for the {\ckd} risk prediction. 
Figure~\ref{fig:shap:summary_proto} (right) shows an alternate view of the same, with further access to the spread of individual importance. From this deeper view, we can see that features such as `calculus of urinary tract' can be the most important drivers of risk for some patients. Such results further support our need for personalized feature importance. 

While such feature importance insights themselves are useful, such clinical and patho-physiological features may need further contextualization for clinicians. 
Annotating higher-level medical concepts would allow the end clinician to formulate a more holistic care plan. To that end, for this POC, we mapped the disease diagnosis features to Level 1 CCS concepts. This grouping approach can be seen in the prototypical patient summary in Table~\ref{tab:protosummary}, where groupings such as `Endocrine, nutritional; and metabolic diseases and immunity disorders,' `Diseases of the circulatory system,' and `Diseases of the  respiratory systems' are identified as the top three most frequent condition groupings among patients in the set. 
Another point to note is that, while some questions can be directly addressed by the application of the guideline QA module (\textbf{Q. 3a, 5, 6}) on recommendations from chapters in the ADA CPG, some others, like \textbf{Q. 4}, need other 
context from the guidelines, such as tabular information. For future work, we plan to extract these data points automatically. 

\vspace{-0.115em}

\section{Methods} \label{sec:methods}
Our POC implementation addresses and enables
the identified contextualizations in the {\ckd} risk prediction among {\dm} patients. This POC is a promising step towards supporting human-in-the-loop interfaces with explanations that clinicians could implement while interacting with their patients.
The POC includes the following:
    \\ \noindent $\bullet$ We select the ML models for predicting the risk based on highest predictive accuracy and other appropriate metrics for the use case, such as favoring models with higher recall. 
    We trained a suite of classification models on the patients' demographic and diagnosis history to predict future complications. The models include both classical such as Logistic Regression (LR) as well as modern deep learning methods  such as Multi-Layer Perceptron (MLP), Long-Short Term Memory (LSTM), and Gated Recurrent Units (GRU). We followed standard ML methods for model evaluation and, based on our results, we selected MLP as our model choice to balance recall with overall performance. For further details, see Appendix~\ref{ssub:risk_model}
    \\ \noindent $\bullet$ We apply Protodash, a post-hoc sample selection method~\cite{gurumoorthy2019efficient}, to select a set of prototypical or representative patients from the high risk category. Prototypical patients from Protodash naturally spans the varied set of patient characteristics for the selected sub-group. This allows the clinician to build trust in using the AI models by inspecting the different patient modalities of the dataset without having to inspect the entire dataset. We also used algorithmic feature importance explainers, such as SHAP~\cite{lundberg2018explainable}, to understand the impact of the clinical factors driving the risk prediction. As explained above, we contextualize such factors by mapping them to medical ontologies to highlight higher-level concepts, such as clincial pathways.
\\ \noindent $\bullet$ We design and utilize a question answering (QA) module to query {\dm}  guideline recommendations, based on patient data. We utilized a language-model approach, applying a BERT~\cite{devlin2018bert} model, pre-trained on the popular Stanford Question Answering Dataset (SQuAD) corpus~\cite{rajpurkar2016squad} on the ADA CPG, to address questions, including details from the patient cohort from recommendations in the {\dm} section of the pharmacological and {\ckd} sections of the microvascular chapters. We validated the answers via expert feedback and plan to conduct quantitative evaluations in the future. Details on the implementation of the QA module are presented in \ref{sec:methods_guidelines}. 

\begin{table}[ht!]
\small
\centering
\caption{Baseline description summary (generated using Tableone library~\cite{pollard2018tableone}) of 20 prototypical patients highlighting the demographic and diagnoses counts. We report the disease diagnoses by their higher-level disease groupings (e.g. for {\dm} the higher-level code is Endocrine, nutritional and metabolic disorders). We present high prevalence conditions 
($>= 50\%$). The entire description is presented in \ref{sec:proto}.}
\label{tab:protosummary}
\begin{tabular}{p{0.78\linewidth}p{0.16\linewidth}}
\toprule
Feature                  &     Count ($\%$) \\
\midrule
n &          20 \\
AGE\_GRP\_M &    4 (20.0) \\
AGE\_GRP\_O &   15 (75.0) \\
AGE\_GRP\_Y &     1 (5.0) \\
SEX - FEMALE &    7 (35.0) \\
Dis. of the circulatory system &   17 (85.0) \\
Dis. of the musculoskeletal system \& connective tissue  &   12 (60.0) \\
Dis. of the respiratory system &   11 (55.0) \\
Endocrine; nutritional; and metabolic dis.  and immunity disorders &  20 (100.0) \\
Infectious and parasitic dis. &   10 (50.0) \\
Symptoms; signs; and ill-defined conditions and factors influencing health status &   10 (50.0) \\
\bottomrule
\end{tabular}
\vspace{-1em}

\end{table}


\vspace{-1.2mm}
\section{Expert Feedback and Discussion}
\label{sec:discussions}
Our work 
was advised and evaluated by a clinical researcher, who is also a clinician by training. Our use case was designed for clinical relevance 
from findings from our past interactions with clinicians~\cite{dan2020designing} and in consultation with the expert. 
The expert opined that our approach holds potential to contextualize and 
connect relevant clinical information, to help clinicians make informed decisions about the entities that they generally interact with when treating patients.
They also evaluated the three dimensions of contextualization we addressed to be of high value. Specifically, the set of questions was evaluated to be of relevance to the clinical workflow. Similarly, the contextualization of the important features, using the medical ontology, followed their advice.

The expert evaluation also 
identifies areas for improvement,
such as increasing the persona coverage beyond PCPs, providing more in-depth contextualization, and supporting more presentation styles~\cite{chari2020explanation} 
including more clinical perspectives.
Our ongoing work, 
is focused on using the patient's context to probe guidelines and create question lists for various types of clinicians at different points in the workflow.
We are also exploring the addition of more 
contextualization dimensions
such as extracting richer content from guidelines and leveraging more biomedical ontologies. We foresee that a contextualization approach like ours can be made applicable to other diseases by utilizing similar post-hoc explainability methods and operationalizing clinical domain knowledge via the identification of relevant ontologies and literature sources.

Overall, the expert's feedback validates our 
goals to make the explanations more adaptive to the end-user's expertise and situate them in the context of the patients they see. 
In continuity, we are building a prototype dashboard and human-centered and qualitative solutions to evaluate the importance of our approach.

\section*{Acknowledgments}
This work is supported by IBM Research AI through the AI Horizons Network. We thank Rebecca Cowan from RPI for their helpful feedback on the work.








\bibliographystyle{ACM-Reference-Format}
\bibliography{references}


\begin{thebibliography}{20}


\ifx \showCODEN    \undefined \def \showCODEN     #1{\unskip}     \fi
\ifx \showDOI      \undefined \def \showDOI       #1{#1}\fi
\ifx \showISBNx    \undefined \def \showISBNx     #1{\unskip}     \fi
\ifx \showISBNxiii \undefined \def \showISBNxiii  #1{\unskip}     \fi
\ifx \showISSN     \undefined \def \showISSN      #1{\unskip}     \fi
\ifx \showLCCN     \undefined \def \showLCCN      #1{\unskip}     \fi
\ifx \shownote     \undefined \def \shownote      #1{#1}          \fi
\ifx \showarticletitle \undefined \def \showarticletitle #1{#1}   \fi
\ifx \showURL      \undefined \def \showURL       {\relax}        \fi
\providecommand\bibfield[2]{#2}
\providecommand\bibinfo[2]{#2}
\providecommand\natexlab[1]{#1}
\providecommand\showeprint[2][]{arXiv:#2}

\bibitem[\protect\citeauthoryear{??}{cdc}{[n.d.]a}]%
        {cdcT2D}
 \bibinfo{year}{[n.d.]}\natexlab{a}.
\newblock \bibinfo{title}{About Chronic Diseases | CDC}.
\newblock
  \bibinfo{howpublished}{\url{https://www.cdc.gov/chronicdisease/about/index.htm}}.
\newblock
\newblock
\shownote{Accessed: 2021-05-26.}


\bibitem[\protect\citeauthoryear{??}{cdc}{[n.d.]b}]%
        {cdcCKD}
 \bibinfo{year}{[n.d.]}\natexlab{b}.
\newblock \bibinfo{title}{Chronic Kidney Disease Basics | Chronic Kidney
  Disease Initiative | CDC}.
\newblock
  \bibinfo{howpublished}{\url{https://www.cdc.gov/kidneydisease/basics.html}}.
\newblock
\newblock
\shownote{Accessed: 2021-05-25.}


\bibitem[\protect\citeauthoryear{Association et~al\mbox{.}}{Association
  et~al\mbox{.}}{2021}]%
        {american2021introduction}
\bibfield{author}{\bibinfo{person}{American~Diabetes Association}
  {et~al\mbox{.}}} \bibinfo{year}{2021}\natexlab{}.
\newblock \bibinfo{title}{Introduction: Standards of Medical Care in
  Diabetes—2021}.
\newblock
\newblock


\bibitem[\protect\citeauthoryear{Chakraborty, Kwon, Dey, Dhurandhar, Gruen, Ng,
  Sow, and Varshney}{Chakraborty et~al\mbox{.}}{2020}]%
        {chakraborty2020tutorial}
\bibfield{author}{\bibinfo{person}{Prithwish Chakraborty},
  \bibinfo{person}{Bum~Chul Kwon}, \bibinfo{person}{Sanjoy Dey},
  \bibinfo{person}{Amit Dhurandhar}, \bibinfo{person}{Daniel Gruen},
  \bibinfo{person}{Kenney Ng}, \bibinfo{person}{Daby Sow}, {and}
  \bibinfo{person}{Kush~R Varshney}.} \bibinfo{year}{2020}\natexlab{}.
\newblock \showarticletitle{Tutorial on Human-Centered Explainability for
  Healthcare}. In \bibinfo{booktitle}{\emph{Proceedings of the 26th ACM SIGKDD
  International Conference on Knowledge Discovery \& Data Mining}}.
  \bibinfo{pages}{3547--3548}.
\newblock


\bibitem[\protect\citeauthoryear{Chari, Seneviratne, Gruen, Foreman, Das, and
  McGuinness}{Chari et~al\mbox{.}}{2020}]%
        {chari2020explanation}
\bibfield{author}{\bibinfo{person}{Shruthi Chari}, \bibinfo{person}{Oshani
  Seneviratne}, \bibinfo{person}{Daniel~M Gruen}, \bibinfo{person}{Morgan~A
  Foreman}, \bibinfo{person}{Amar~K Das}, {and} \bibinfo{person}{Deborah~L
  McGuinness}.} \bibinfo{year}{2020}\natexlab{}.
\newblock \showarticletitle{Explanation Ontology: A Model of Explanations for
  User-Centered AI}. In \bibinfo{booktitle}{\emph{International Semantic Web
  Conference}}. Springer, \bibinfo{pages}{228--243}.
\newblock


\bibitem[\protect\citeauthoryear{Chen, Subburathinam, Chen, and Zaki}{Chen
  et~al\mbox{.}}{2021}]%
        {chen2021personalized}
\bibfield{author}{\bibinfo{person}{Yu Chen}, \bibinfo{person}{Ananya
  Subburathinam}, \bibinfo{person}{Ching-Hua Chen}, {and}
  \bibinfo{person}{Mohammed~J Zaki}.} \bibinfo{year}{2021}\natexlab{}.
\newblock \showarticletitle{Personalized Food Recommendation as Constrained
  Question Answering over a Large-scale Food Knowledge Graph}.
\newblock \bibinfo{journal}{\emph{arXiv preprint arXiv:2101.01775}}
  (\bibinfo{year}{2021}).
\newblock


\bibitem[\protect\citeauthoryear{Devlin, Chang, Lee, and Toutanova}{Devlin
  et~al\mbox{.}}{2018}]%
        {devlin2018bert}
\bibfield{author}{\bibinfo{person}{Jacob Devlin}, \bibinfo{person}{Ming-Wei
  Chang}, \bibinfo{person}{Kenton Lee}, {and} \bibinfo{person}{Kristina
  Toutanova}.} \bibinfo{year}{2018}\natexlab{}.
\newblock \showarticletitle{Bert: Pre-training of deep bidirectional
  transformers for language understanding}.
\newblock \bibinfo{journal}{\emph{arXiv preprint arXiv:1810.04805}}
  (\bibinfo{year}{2018}).
\newblock


\bibitem[\protect\citeauthoryear{Doshi-Velez, Kortz, Budish, Bavitz, Gershman,
  O'Brien, Scott, Schieber, Waldo, Weinberger, et~al\mbox{.}}{Doshi-Velez
  et~al\mbox{.}}{2017}]%
        {doshi2017accountability}
\bibfield{author}{\bibinfo{person}{Finale Doshi-Velez}, \bibinfo{person}{Mason
  Kortz}, \bibinfo{person}{Ryan Budish}, \bibinfo{person}{Chris Bavitz},
  \bibinfo{person}{Sam Gershman}, \bibinfo{person}{David O'Brien},
  \bibinfo{person}{Kate Scott}, \bibinfo{person}{Stuart Schieber},
  \bibinfo{person}{James Waldo}, \bibinfo{person}{David Weinberger},
  {et~al\mbox{.}}} \bibinfo{year}{2017}\natexlab{}.
\newblock \showarticletitle{Accountability of AI under the law: The role of
  explanation}.
\newblock \bibinfo{journal}{\emph{arXiv preprint arXiv:1711.01134}}
  (\bibinfo{year}{2017}).
\newblock


\bibitem[\protect\citeauthoryear{Gruen, Chari, , Foreman, Seneviratne,
  Richesson, Das, and McGuinness}{Gruen et~al\mbox{.}}{2021}]%
        {dan2020designing}
\bibfield{author}{\bibinfo{person}{Daniel~M Gruen}, \bibinfo{person}{Shruthi
  Chari}, \bibinfo{person}{}, \bibinfo{person}{Morgan~A Foreman},
  \bibinfo{person}{Oshani Seneviratne}, \bibinfo{person}{Rachel Richesson},
  \bibinfo{person}{Amar~K Das}, {and} \bibinfo{person}{Deborah~L McGuinness}.}
  \bibinfo{year}{2021}\natexlab{}.
\newblock \showarticletitle{Designing for AI Explainability in Clinical
  Context}. In \bibinfo{booktitle}{\emph{Trustworthy AI for Healthcare Workshop
  at AAAI 2021}}.
\newblock


\bibitem[\protect\citeauthoryear{Gurumoorthy, Dhurandhar, Cecchi, and
  Aggarwal}{Gurumoorthy et~al\mbox{.}}{2019}]%
        {gurumoorthy2019efficient}
\bibfield{author}{\bibinfo{person}{Karthik~S Gurumoorthy},
  \bibinfo{person}{Amit Dhurandhar}, \bibinfo{person}{Guillermo Cecchi}, {and}
  \bibinfo{person}{Charu Aggarwal}.} \bibinfo{year}{2019}\natexlab{}.
\newblock \showarticletitle{Efficient data representation by selecting
  prototypes with importance weights}. In \bibinfo{booktitle}{\emph{2019 IEEE
  International Conference on Data Mining (ICDM)}}. IEEE,
  \bibinfo{pages}{260--269}.
\newblock


\bibitem[\protect\citeauthoryear{Kwon, Chakraborty, Codella, Dhurandhar, Sow,
  and Ng}{Kwon et~al\mbox{.}}{2020}]%
        {up:bckwon2020visual}
\bibfield{author}{\bibinfo{person}{Bum~Chul Kwon}, \bibinfo{person}{Prithwish
  Chakraborty}, \bibinfo{person}{James Codella}, \bibinfo{person}{Amit
  Dhurandhar}, \bibinfo{person}{Daby Sow}, {and} \bibinfo{person}{Kenney Ng}.}
  \bibinfo{year}{2020}\natexlab{}.
\newblock \bibinfo{title}{{Visually Exploring Contrastive Explanation for
  Diagnostic Risk Prediction on Electronic Health Records}}.
\newblock \bibinfo{howpublished}{\url{http://whi2020.online/poster_76.html}}.
\newblock


\bibitem[\protect\citeauthoryear{Lundberg, Nair, Vavilala, Horibe, Eisses,
  Adams, Liston, Low, Newman, Kim, et~al\mbox{.}}{Lundberg
  et~al\mbox{.}}{2018}]%
        {lundberg2018explainable}
\bibfield{author}{\bibinfo{person}{Scott~M Lundberg}, \bibinfo{person}{Bala
  Nair}, \bibinfo{person}{Monica~S Vavilala}, \bibinfo{person}{Mayumi Horibe},
  \bibinfo{person}{Michael~J Eisses}, \bibinfo{person}{Trevor Adams},
  \bibinfo{person}{David~E Liston}, \bibinfo{person}{Daniel King-Wai Low},
  \bibinfo{person}{Shu-Fang Newman}, \bibinfo{person}{Jerry Kim},
  {et~al\mbox{.}}} \bibinfo{year}{2018}\natexlab{}.
\newblock \showarticletitle{Explainable machine-learning predictions for the
  prevention of hypoxaemia during surgery}.
\newblock \bibinfo{journal}{\emph{Nature Biomedical Engineering}}
  \bibinfo{volume}{2}, \bibinfo{number}{10} (\bibinfo{year}{2018}),
  \bibinfo{pages}{749}.
\newblock


\bibitem[\protect\citeauthoryear{Matheny, Thadaney, Ahmed, and Whicher}{Matheny
  et~al\mbox{.}}{tion}]%
        {namreport2019}
\bibfield{author}{\bibinfo{person}{ME Matheny}, \bibinfo{person}{Israni~S
  Thadaney}, \bibinfo{person}{M Ahmed}, {and} \bibinfo{person}{D. Whicher}.}
  \bibinfo{year}{2019. [Online] Available:
  \url{https://nam.edu/artificial-intelligence-special-publication}}\natexlab{}.
\newblock \showarticletitle{Artificial Intelligence in Health Care: The Hope,
  the Hype, the Promise, the Peril}.
\newblock \bibinfo{journal}{\emph{National Academy of Medicine}}
  (\bibinfo{year}{2019. [Online] Available:
  \url{https://nam.edu/artificial-intelligence-special-publication}}).
\newblock


\bibitem[\protect\citeauthoryear{Mittelstadt, Russell, and Wachter}{Mittelstadt
  et~al\mbox{.}}{2019}]%
        {mittelstadt2019explaining}
\bibfield{author}{\bibinfo{person}{Brent Mittelstadt}, \bibinfo{person}{Chris
  Russell}, {and} \bibinfo{person}{Sandra Wachter}.}
  \bibinfo{year}{2019}\natexlab{}.
\newblock \showarticletitle{Explaining explanations in AI}. In
  \bibinfo{booktitle}{\emph{Proceedings of the conference on fairness,
  accountability, and transparency}}. ACM, \bibinfo{pages}{279--288}.
\newblock


\bibitem[\protect\citeauthoryear{Pollard, Johnson, Raffa, and Mark}{Pollard
  et~al\mbox{.}}{2018}]%
        {pollard2018tableone}
\bibfield{author}{\bibinfo{person}{Tom~J Pollard}, \bibinfo{person}{Alistair~EW
  Johnson}, \bibinfo{person}{Jesse~D Raffa}, {and} \bibinfo{person}{Roger~G
  Mark}.} \bibinfo{year}{2018}\natexlab{}.
\newblock \showarticletitle{tableone: An open source Python package for
  producing summary statistics for research papers}.
\newblock \bibinfo{journal}{\emph{JAMIA open}} \bibinfo{volume}{1},
  \bibinfo{number}{1} (\bibinfo{year}{2018}), \bibinfo{pages}{26--31}.
\newblock


\bibitem[\protect\citeauthoryear{Raghu, Guttag, Young, Pomerantsev, Dalca, and
  Stultz}{Raghu et~al\mbox{.}}{2021}]%
        {raghu2021learning}
\bibfield{author}{\bibinfo{person}{Aniruddh Raghu}, \bibinfo{person}{John
  Guttag}, \bibinfo{person}{Katherine Young}, \bibinfo{person}{Eugene
  Pomerantsev}, \bibinfo{person}{Adrian~V Dalca}, {and}
  \bibinfo{person}{Collin~M Stultz}.} \bibinfo{year}{2021}\natexlab{}.
\newblock \showarticletitle{Learning to predict with supporting evidence:
  applications to clinical risk prediction}.
\newblock  (\bibinfo{year}{2021}), \bibinfo{pages}{95--104}.
\newblock


\bibitem[\protect\citeauthoryear{Rajpurkar, Zhang, Lopyrev, and
  Liang}{Rajpurkar et~al\mbox{.}}{2016}]%
        {rajpurkar2016squad}
\bibfield{author}{\bibinfo{person}{Pranav Rajpurkar}, \bibinfo{person}{Jian
  Zhang}, \bibinfo{person}{Konstantin Lopyrev}, {and} \bibinfo{person}{Percy
  Liang}.} \bibinfo{year}{2016}\natexlab{}.
\newblock \showarticletitle{Squad: 100,000+ questions for machine comprehension
  of text}.
\newblock \bibinfo{journal}{\emph{arXiv preprint arXiv:1606.05250}}
  (\bibinfo{year}{2016}).
\newblock


\bibitem[\protect\citeauthoryear{Seroussi, Lamy, Muro, Larburu, Sekar,
  Gu{\'e}zennec, and Bouaud}{Seroussi et~al\mbox{.}}{2018}]%
        {seroussi2018implementing}
\bibfield{author}{\bibinfo{person}{Brigitte Seroussi},
  \bibinfo{person}{Jean-Baptiste Lamy}, \bibinfo{person}{Naiara Muro},
  \bibinfo{person}{Nekane Larburu}, \bibinfo{person}{Booma~Devi Sekar},
  \bibinfo{person}{Gilles Gu{\'e}zennec}, {and} \bibinfo{person}{Jacques
  Bouaud}.} \bibinfo{year}{2018}\natexlab{}.
\newblock \showarticletitle{Implementing Guideline-Based, Experience-Based, and
  Case-Based Approaches to Enrich Decision Support for the Management of Breast
  Cancer Patients in the DESIREE Project.}. In
  \bibinfo{booktitle}{\emph{EFMI-STC}}. \bibinfo{pages}{190--194}.
\newblock


\bibitem[\protect\citeauthoryear{Shortliffe}{Shortliffe}{1974}]%
        {shortliffe1974mycin}
\bibfield{author}{\bibinfo{person}{Edward~Hance Shortliffe}.}
  \bibinfo{year}{1974}\natexlab{}.
\newblock \bibinfo{booktitle}{\emph{MYCIN: a rule-based computer program for
  advising physicians regarding antimicrobial therapy selection.}}
\newblock \bibinfo{type}{{T}echnical {R}eport}. \bibinfo{institution}{Dept. of
  Computer Sci., Stanford University Stanford}.
\newblock


\bibitem[\protect\citeauthoryear{Tonekaboni, Joshi, McCradden, and
  Goldenberg}{Tonekaboni et~al\mbox{.}}{2019}]%
        {tonekaboni2019clinicians}
\bibfield{author}{\bibinfo{person}{Sana Tonekaboni}, \bibinfo{person}{Shalmali
  Joshi}, \bibinfo{person}{Melissa~D McCradden}, {and} \bibinfo{person}{Anna
  Goldenberg}.} \bibinfo{year}{2019}\natexlab{}.
\newblock \showarticletitle{What clinicians want: contextualizing explainable
  machine learning for clinical end use}. In \bibinfo{booktitle}{\emph{Machine
  Learning for Healthcare Conference}}. PMLR, \bibinfo{pages}{359--380}.
\newblock


\end{thebibliography}

\appendix

\section{Research Methods}
\subsection{Architecture}
\begin{figure*}
  \includegraphics[width=\textwidth]{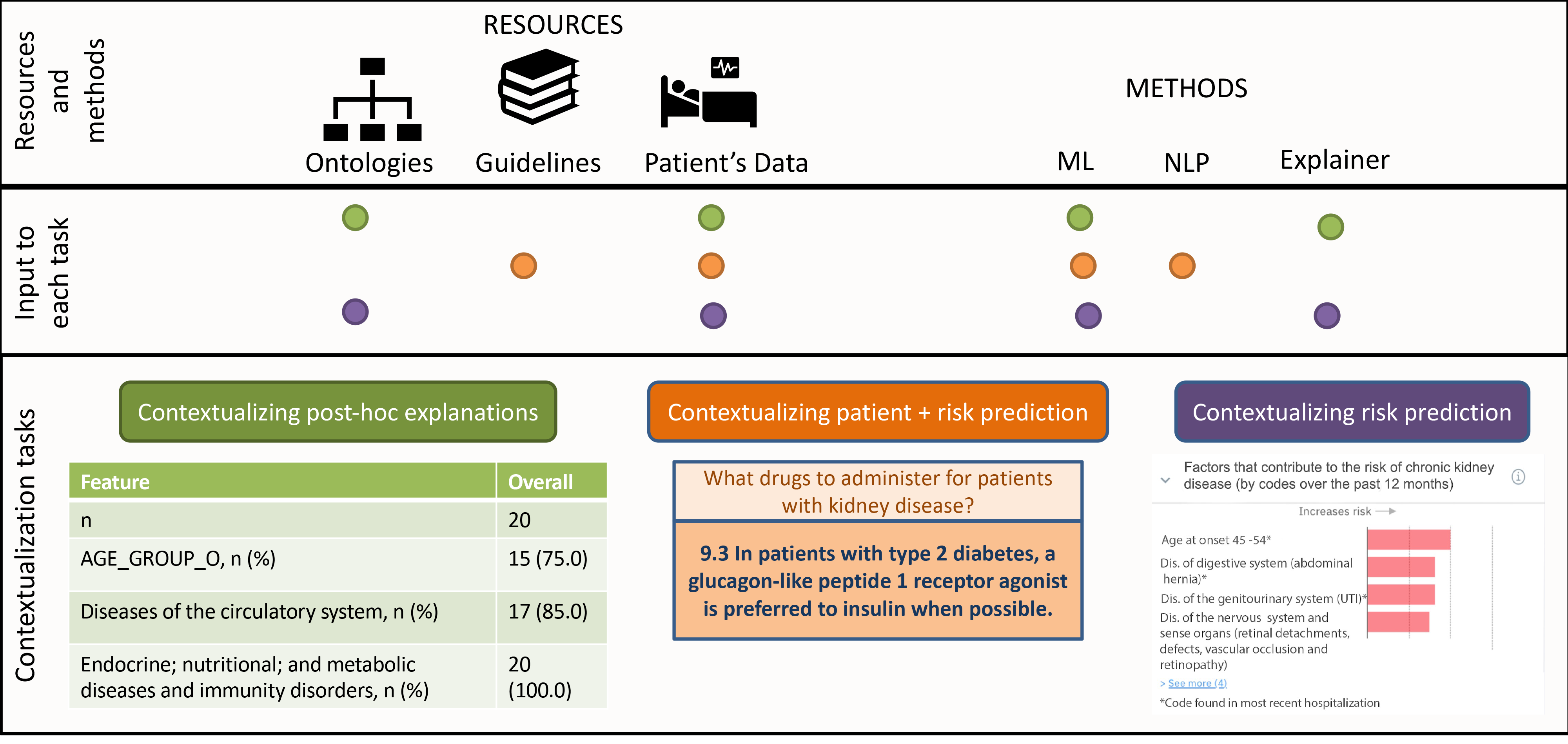}
  \caption{An overview of the interactions between different methods in the pipeline to generate insights towards the three identified contextualization capabilities of contextualizing a patient, their risk prediction and post-hoc explanations. We demonstrate the different parts used by the contextualizations, shown to the right of the figure, via coloured swimlane paths. }
  \label{fig:teaser}
\end{figure*}

An architectural view of our POC implementation is 
shown in Fig. \ref{fig:teaser}.

\subsection{Risk Prediction Model}
\label{ssub:risk_model}
To model the risk of the complications, we used a number of classical as well as deep-learning models. For each patient, we have a temporal history of their demographic, diagnostic, drug, and clinical features. In this paper, we only used the demographic and diagnostic features to model the risk. Furthermore, to handle the temporal features, for some of our models, such as Logistic Regression (LR) and Multi-layer perceptron (MLP), we used temporally aggregated features (summation). 
We also compared two state-of-the art Recurrent Neural Networks (RNN) where temporal history can be handled in a more natural manner viz. Long-short term memory (LSTM) and Gated Recurrent Units (GRU).
In this paper, we split the data according to a train-validation-test split (70-10-20)
and present our results on the hold-out test set for the best performing models on the validation set. Since the data is imbalanced, we selected the models based on the best AUC-ROC and AUC-PRC from the validation set. We also evaluate the models based on precision, recall, and brier score. Deep learning networks are known to be under-calibrated and the last metric measures how well the model is calibrated i.e. it measures the probabilistic interpretation of the risk prediction. In other words, if a model predicts a $0.7$ risk for a patient, brier score measures whether that translates to a $70\%$ chance of the patient developing the complication.

\begin{table}
  \caption{Results of Risk Prediction Model for {\ckd} 
  }
  \label{tab:resultsCKD}
  \begin{tabular}{lccccc}
    \toprule
    Method & Precision & Recall & AUC-ROC & AUC-PRC & Brier\\
    \midrule
    LR    &  0.333 & 0.023  & 0.582 & 0.215 &  0.127 \\
    MLP   &  0.139 &  {\bf 0.977} & 0.587 & 0.224 & {\bf 0.621}  \\
    LSTM  &  {\bf 0.242} &  0.442 & {\bf 0.678} & 0.263 & 0.208  \\
    GRU   &  0.240  &  {\bf 0.605} & 0.677 & {\bf 0.311} & 0.220   \\
  \bottomrule
\end{tabular}
\end{table}

We present the performance of the risk prediction models in Table~\ref{tab:resultsCKD}. As can be seen, while GRU performs the best overall, depending on the use case we may want to prefer other models. For the purposes of this paper, we chose MLP as our risk prediction model to benefit from the higher recall (such that the probability of false negatives is low) and high brier-score (to allow a more natural interpretation of our model outputs to clinicians) while still achieving an acceptable level of overall performance (AUC-ROC = $0.59$).

\subsection{Post-hoc Explainers} \label{sec:pec}
While some of the classical algorithms considered in Section~\ref{ssub:risk_model} are inherently interpretable with easy access to the features deemed important for the model (such as LR), several of the deep learning models are black-box models. 
To extract feature importances from such models, we used post-hoc explainers which have been found to be favored by clinicians in past studies~\cite{tonekaboni2019clinicians}.
In particular, we used the well accepted SHAP algorithm~\cite{lundberg2018explainable} to find feature importance. The algorithm uses game-theoretic principles to identify importance of features by ascertaining the dip in performance of the model with and without access to the feature at the personalized level. 
Such personalized feature importance is key so that our overall dashboards are more actionable for the clinicians by allowing them to focus on the particular attributes of the patients that are driving their risk.
However, for a user study, it is impractical for a clinician to review the model predictions for all the patients in the test set. Thus we follow the principles outlined in~\cite{up:bckwon2020visual} to select a few representative (or prototypical) patients from the feature set using a self-supervised algorithm called Protodash~\cite{gurumoorthy2019efficient} and analyze the feature importance on these selected set of patients. This also allows our clincial experts to review a diverse and representative set of patients among the test set and provide feedback in a more practical manner while reducing selection bias of patients compared to random draw of patients.

\subsection{Selection of prototypical patients} \label{sec:proto}
Recently, a significant amount of research
has been focused on summarizing datasets based on representative or prototypical 
examples. \cite{gurumoorthy2019efficient} introduced `ProtoDash' 
that can find such prototypical examples that best summarizes and compactly represents
the underlying data distribution of a population. `ProtoDash' can find such prototypes
along with non-negative importance weights of the instances that allows the user
to better comprehend the samples. Furthermore, it can be used to find both representative
samples (prototypes) as well as outliers (criticisms) from the population dataset.
In this work, we use ProtoDash to find a small number of representative samples 
that are then investigated for local explainability using post-hoc explanations (see Table \ref{tab:protosummary_appendix}).

\vspace{-1mm}
\begin{table}
\caption{Entire baseline description summary (generated using Tableone library~\cite{pollard2018tableone}) of the 20 prototypical patients highlighting the demographic and diagnoses counts. We report the disease diagnoses by their higher-level disease groupings (e.g. for {\dm} the higher-level code is Endocrine, nutritional and metabolic disorders). We highlight the conditions that are most prevalent amongst the patients ($> 50\%$).}
\label{tab:protosummary_appendix}
 \begin{tabular}{lcc}
\toprule
                                                                       Feature                  &     Overall \\
\midrule
n &          20 \\
AGE\_GRP\_M, n (\%) &    4 (20.0) \\
\textbf{AGE\_GRP\_O, n (\%)} &   \textbf{15 (75.0)} \\
AGE\_GRP\_Y, n (\%) &     1 (5.0) \\
SEX - FEMALE, n (\%) &    7 (35.0) \\
Mood disorders, n (\%)  &    3 (15.0) \\
Diseases of the blood and blood-forming organs, n (\%)  &    3 (15.0) \\
\textbf{Diseases of the circulatory system, n (\%)} &   \textbf{17 (85.0)} \\
Diseases of the digestive system, n (\%) &    6 (30.0) \\
Diseases of the genitourinary system, n (\%) &    9 (45.0) \\
\begin{tabular}[c]{@{}l@{}} \textbf{Diseases of the musculoskeletal system} \\ \textbf{and connective tissue, n (\%)} \end{tabular}  &   \textbf{12 (60.0)} \\
Diseases of the nervous system and sense organs, n (\%)  &    9 (45.0) \\
\textbf{Diseases of the respiratory system, n (\%)} &   \textbf{11 (55.0)} \\
Diseases of the skin and subcutaneous tissue, n (\%) &    7 (35.0) \\
\begin{tabular}[c]{@{}l@{}} \textbf{Endocrine; nutritional; and metabolic} \\ \textbf{ diseases and immunity disorders, n (\%)} \end{tabular} &  \textbf{20 (100.0)} \\
Infectious and parasitic diseases, n (\%) &   10 (50.0) \\
Injury and poisoning, n (\%) &    4 (20.0) \\
Mental Illness, n (\%)  &    3 (15.0) \\
Neoplasms, n (\%)  &    6 (30.0) \\
\begin{tabular}[c]{@{}l@{}} Symptoms; signs; and ill-defined conditions \\ and factors influencing health status, n (\%) \end{tabular} &   10 (50.0) \\
\bottomrule
\end{tabular}
\end{table}

\subsection{Guideline Question Answering} \label{sec:methods_guidelines}

\begin{figure}[th!]
  \centering
  \includegraphics[width=1.00\linewidth]{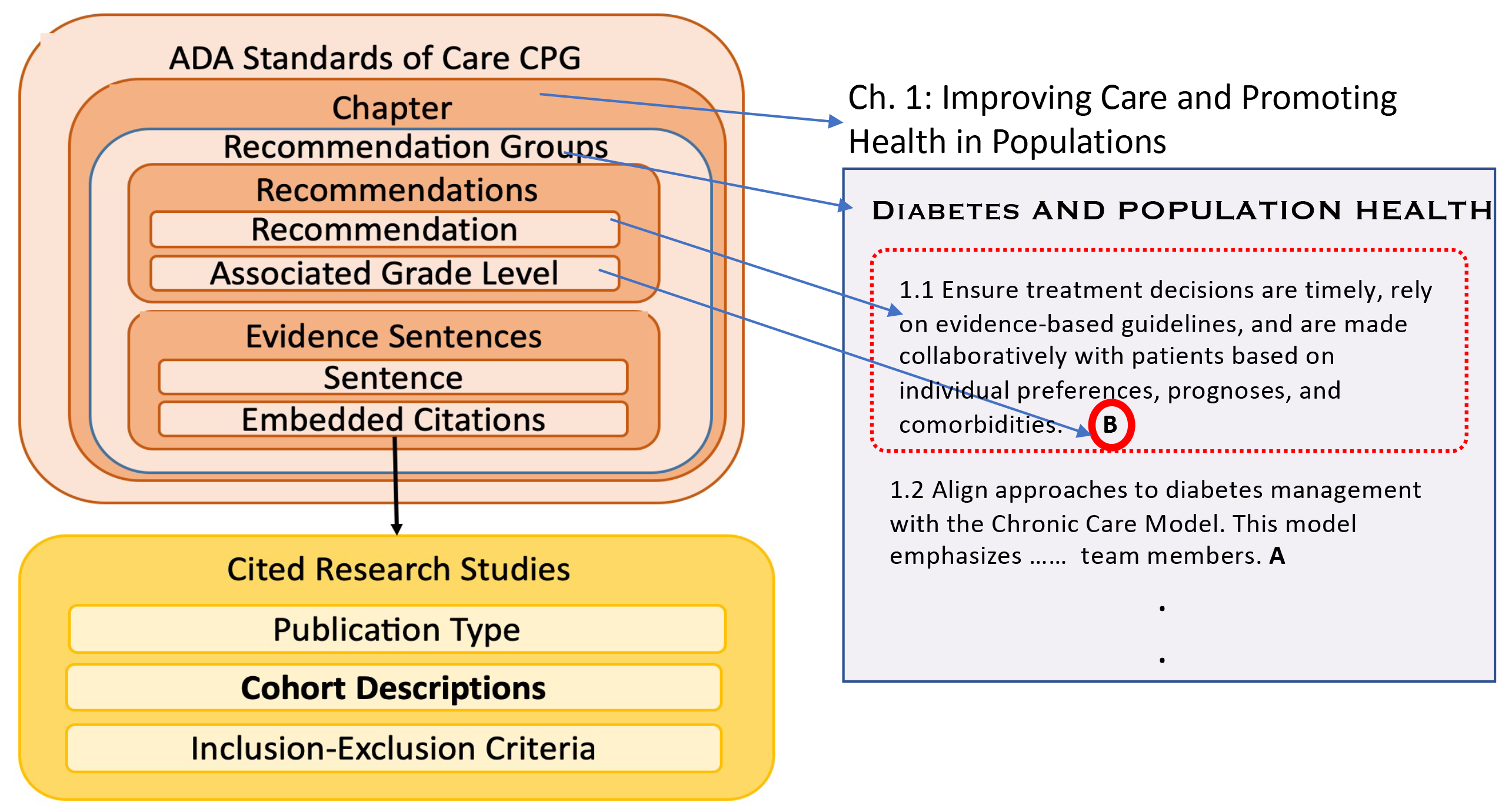}
   \caption{ADA CPG evidence structure with its compositional structure.  Chapters contain recommendation groups covering sub-topics that further contain recommendations that have grade levels dictated by the evidence quality.}
  \label{fig:guideline}
\end{figure}

To utilize the ADA guidelines in a question-answer setup, we extract content from the guidelines in a JSON structure for machine-usable rendering, and then we apply a language model approach to locate the most probable answer for a question. The guidelines are largely made up of text snippets, so we decided to use a flexible language model approach to enable question answering (QA) capabilities. As an initial focus, we feed the recommendation natural-language text alone (as can be seen in Fig. \ref{fig:guideline}) to a pre-trained BERT-based~\cite{devlin2018bert} QA module trained on the widely-used Stanford Question Answering Dataset (SQuAD) corpus~\cite{rajpurkar2016squad}. The BERT-based QA module returns the maximum length matched token for each question, and we retrieve the matched guideline recommendation for this returned token.

Further, the BERT-based QA module could not handle questions with numerical ranges, so inspired by the Chen et al.~\cite{chen2021personalized} approach, we wrote a parsing module that identifies numerical phrases in both the question and answer and checks to see if the numerical ranges in the question lie within that of the answer. This numerical range feature helps answer questions involving a patient's lab parameters such as `Find recommendations for A1C levels greater than 10'. From our assessment, we find that the current BERT model can handle different questions that utilize parameters from the patient cohort, as seen in Table \ref{tab:resultsquestions}. In this manner, we set up the guideline content to be used as domain knowledge to address questions about the {\dm} patient (s) or their risk prediction for {\ckd}. 








\end{document}